%% file: main.tex
\documentclass[runningheads]{llncs}

\usepackage{eccv}

\usepackage{eccvabbrv}
\usepackage{fontawesome}
\usepackage{graphicx}
\usepackage{booktabs}
\usepackage{algorithm}
\usepackage{algpseudocode}
\usepackage{makecell}

\usepackage{footmisc}

\newcommand\blfootnote[1]{%
  \begingroup
  \renewcommand\thefootnote{}%
  \begin{NoHyper}%
  \footnotetext{\hspace{-0.5em}#1}%
  \end{NoHyper}%
  \endgroup
}
\usepackage[accsupp]{axessibility}  
\usepackage{multirow}
\usepackage[table]{xcolor}
\usepackage{bm}
\usepackage{wrapfig}

\usepackage{hyperref}
\usepackage{orcidlink}

\begin{document}

\title{MindDrive: A Vision-Language-Action Model for Autonomous Driving via Online Reinforcement Learning} 



\author{Haoyu Fu$^{1*}$, Diankun Zhang$^{2*}$, Zongchuang Zhao$^{1}$, Jianfeng Cui$^{2}$, \\ Hongwei Xie$^{2\dag}$, Bing Wang$^{2}$, Guang Chen$^{2}$, Hangjun Ye$^{2}$, \\ Dingkang Liang$^{1}$\textsuperscript{\faEnvelopeO}, Xiang Bai$^{1}$ \\
{\tt \{hyfu, zcuangzhao, dkliang\}@hust.edu.cn} 
\url{https://xiaomi-mlab.github.io/MindDrive/}
}
\authorrunning{H. Fu et al.}

\institute{Huazhong University of Science and Technology, China \and 
Xiaomi EV, China\\}

\maketitle

\blfootnote{
  * Equal contribution. \quad $\dagger$ Project leader. \quad 
  \texorpdfstring{\faEnvelopeO}{}~Corresponding author. 
  Work done when Haoyu Fu was an intern at Xiaomi EV.
}

\begin{abstract}
Current Vision-Language-Action (VLA) paradigms in autonomous driving primarily rely on Imitation Learning (IL), which introduces inherent challenges such as distribution shift and causal confusion. Online Reinforcement Learning offers a promising pathway to address these issues through trial-and-error learning. However, applying online reinforcement learning to VLA models in autonomous driving is hindered by inefficient exploration in continuous action spaces. To overcome this limitation, we propose MindDrive, a VLA framework comprising a large language model (LLM) with two distinct sets of LoRA parameters. The one LLM serves as a Decision Expert for scenario reasoning and driving decision-making, while the other acts as an Action Expert that dynamically maps linguistic decisions into feasible trajectories. By feeding trajectory-level rewards back into the reasoning space, MindDrive enables trial-and-error learning over a finite set of discrete linguistic driving decisions, instead of operating directly in a continuous action space. This approach effectively balances optimal decision-making in complex scenarios, human-like driving behavior, and efficient exploration in online reinforcement learning. Extensive experiments validate the efficacy of our online reinforcement learning framework, which outperforms state-of-the-art IL and offline Reinforcement Learning methods on the challenging Bench2Drive benchmark. To the best of our knowledge, this is the first work to demonstrate the effective application of online reinforcement learning to VLA models in autonomous driving.

\end{abstract}

\section{Introduction}
\label{sec:intro}
Autonomous driving relies on the model's ability to perceive, make decisions, and act in dynamic complex environments. Traditional end-to-end autonomous driving frameworks~\cite{hu2023planning,jiang2023vad,zhang2024sparsead} integrate perception, prediction, and planning modules, but they lack common sense and causal reasoning abilities. With the improved visual understanding and reasoning ability of Vision Language Models (VLMs)~\cite{openai2023gpt, anil2023gemini, wang2024qwen2vl}, many research~\cite{xu2025drivegpt4,renz2025simlingo,fu2025orion} attempts to apply the Vision-Language-Action (VLA) paradigm to the field of end-to-end autonomous driving. The VLA paradigm in autonomous driving aims to translate complex traffic scene understanding into trajectories of the ego vehicle.

Current VLA models~\cite{fu2025orion,xu2024drivegpt4,hwang2024emma} are primarily trained using imitation learning (IL), which aims to fit expert behaviors from collected driving data. However, exclusive reliance on the IL paradigm renders models susceptible to causal confusion and distributional shift~\cite{de2019causal,ross2011reduction},  resulting in irreversible error accumulation in closed-loop driving scenarios. Reinforcement learning offers a perspective for addressing these challenges through trial-and-error and has achieved notable success in enhancing the causal reasoning capabilities of VLMs~\cite{guo2025deepseek,rafailov2023direct,ouyang2022training}.

\begin{figure}[t!]
    \centering
    \includegraphics[width=0.99\textwidth]{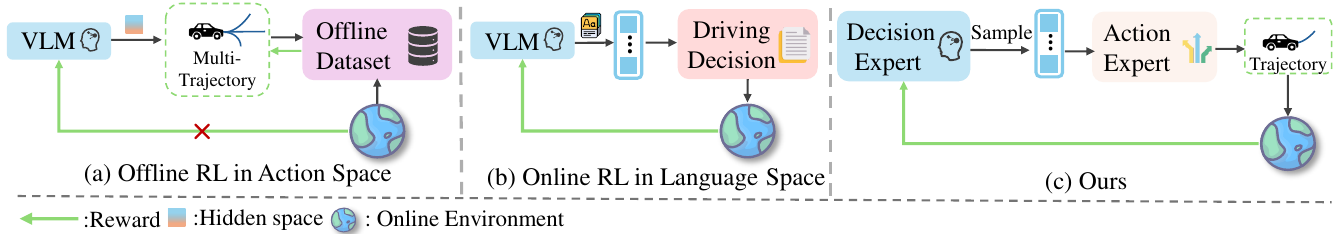}
    \caption{The comparison of different VLA for reinforcement learning paradigms. The proposed Decision Expert and Action Expert share the same base LLM with distinct LoRA adapters.}
    \label{fig: introduction}
\end{figure}

Unlike reinforcement learning in VLMs' discrete language space, the action space of autonomous driving is a continuous trajectory space. Applications of reinforcement learning in the current VLA domain for autonomous driving can be categorized into two main paradigms: offline reinforcement learning for the action space and online reinforcement learning for the language space. Offline reinforcement learning is typically trained on fixed datasets composed of expert demonstrations, as shown in Fig.~\ref{fig: introduction} (a). These methods~\cite{zhou2025autovla, li2025recogdrive,yuan2025autodrive,jiang2025irl} employ offline reinforcement learning with distinct reward functions~\cite{dauner2024navsim} to generate more feasible trajectories in the action space. Although they made great progress, offline reinforcement learning limits the VLA model's ability to explore the environment through interaction. Additionally, trajectory optimization in reinforcement learning cannot effectively improve VLMs' reasoning ability. To overcome these limitations, some other methods~\cite{huang2025vlm, jiang2025alphadrive, Zhang_2026_CVPR} attempt to employ online reinforcement learning in the language space, as shown in Fig.~\ref{fig: introduction} (b). These methods formulate autonomous driving tasks as VQA tasks and treat driving decisions as actions, thereby facilitating deeper causal reasoning via online reinforcement learning. However, they struggle to map driving decisions to specific, human-like driving trajectories. Therefore, utilizing online reinforcement learning to enhance the performance of autonomous driving VLA requires further exploration. 

To address these challenges, we propose a novel architecture, MindDrive, a VLA model for autonomous driving via
online reinforcement learning, as shown in Fig.~\ref{fig: introduction} (c). MindDrive transforms the action space from  continuous trajectories into discrete language-based decisions by dynamically mapping, significantly improving exploration efficiency while using trajectory rewards to reinforce the model's reasoning in online reinforcement learning. Specifically, MindDrive consists of two experts initialized from the same LLM and differing only in their respective LoRA adapters~\cite{hu2021lora}. One LLM acts as a Decision Expert, responsible for making reasonable decisions based on the current scenario, while the other serves as an Action Expert, establishing a dynamic mapping from the reasoning result to continuous trajectories. MindDrive first leverages IL to establish a one-to-one correspondence between the meta-actions inferred by the Decision Expert and the multi-modal trajectories output by the Action Expert. The high-quality driving trajectories generated by the Action Expert provide reasonable, human-style candidates for online reinforcement learning. Subsequently, we leverage action trajectory feedback within an online reinforcement learning framework to refine the Decision Expert, allowing it to learn optimal policies by sampling diverse trajectories and receiving corresponding rewards from the interactive environment.


Meanwhile, to enable model exploration and training in a dynamic, interactive environment, we introduce an online closed-loop reinforcement learning framework for autonomous driving VLA models, built upon the CARLA simulator~\cite{dosovitskiy2017carla}. We define explicit signals for task success and failure, and partition the online reinforcement learning process into data-collection and training phases. During collection, we compute and cache scene tokens for each frame, which serve as compact state representations. This pre-computation step reduces memory buffer overhead, enables large-batch training, and allows the entire process to be formulated as a standard Markov Decision Process~\cite{puterman1990markov}.

We evaluate the driving ability of MindDrive on the comprehensive and challenging closed-loop benchmark, Bench2Drive~\cite{jia2024bench2drive}. Extensive experiments demonstrate that our framework leads to more effective driving behavior in complex driving scenarios. Employing a lightweight 0.5B LLM~\cite{qwen2}, MindDrive outperforms the strong baseline~\cite{fu2025orion} with the same parameter size by 5.15\% in Driving Score (DS) and 9.26\% in Success Rate (SR), respectively. Furthermore, with the 3B LLM~\cite{qwen2.5}, MindDrive achieves 80.59 DS and 58.26\% SR, significantly boosting performance and setting a new state-of-the-art (SOTA) on the benchmark.

The main contributions of this paper are as follows:
\begin{itemize}
\item We propose MindDrive, an online reinforcement learning framework for VLA autonomous driving models. By introducing a dynamic language-action mapping, MindDrive significantly improves exploration efficiency and enables trajectory-level action rewards to facilitate reasoning optimization.

\item We introduce an online reinforcement learning scheme. To the best of our knowledge, MindDrive is the first VLA-based autonomous driving model trained with online reinforcement learning in a simulator, aiming to bring new inspiration to the autonomous driving community.
\item Extensive experiments demonstrate the effectiveness of MindDrive, achieving 80.59 DS and 58.26\% SR on Bench2Drive benchmark, significantly surpassing SOTA IL and reinforcement learning baselines at the same scale. 

\end{itemize}

\section{Related work}

\subsection{End-to-End Autonomous Driving (AD)}

End-to-end (E2E) autonomous driving (AD) based on imitation learning~\cite{pan2018agile, jia2024amp, hu2023planning, jiang2023vad} integrates perception, prediction, and planning into a differentiable framework~\cite{zhang2023fully, mao2024pillarnest, cheng2024rethinking, liu2024laformer}. These models directly map sensor inputs to trajectories or control commands from expert data. Representative methods such as UniAD~\cite{hu2023planning} and VAD~\cite{jiang2023vad} adopt dense bird’s-eye view (BEV) representations to unify multiple tasks. Generative approaches, including GenAD~\cite{zheng2024genad} and DiffusionDrive~\cite{liao2025diffusiondrive}, employ diffusion-based planning for multi-modal trajectory generation. DiffAD~\cite{wang2025diffad} rasterizes heterogeneous driving targets into a unified BEV space for joint optimization. SparseAD~\cite{zhang2024sparsead} and DriveTransformer~\cite{jiadrivetransformer} improve efficiency with sparse queries.


Despite strong open-loop results, E2E models often fail in closed-loop due to distribution shift and causal confusion~\cite{caesar2020nuscenes, jia2024bench2drive}. To overcome these limitations, recent studies have explored integrating vision-language models (VLMs) for AD, leveraging their strong generalization, reasoning, and world knowledge~\cite{wu2026generation, guan2026video}. HERMES~\cite{zhou2025hermes, zhou2026hermespp} proposes a unified model of the AD world that integrates 3D scene understanding and future generation for the first time. DriveGPT4~\cite{xu2024drivegpt4}, ELM~\cite{zhou2024embodied}, and DriveMM~\cite{huang2025robotron} incorporate VLMs for perception, prediction, and decision-making.  OmniDrive~\cite{wang2025omnidrive}, EMMA~\cite{hwang2024emma}, and OpenEMMA~\cite{xing2025openemma} represent trajectories as text with chain-of-thought reasoning~\cite{wei2022chain}. AutoVLA~\cite{zhou2025autovla} discretizes actions for alignment with language models but struggles with continuous motion. SimLingo~\cite{renz2025simlingo} proposes the action dreaming task to achieve alignment between language instructions and driving actions.

Vision-language-action (VLA) models from embodied AI, such as OpenVLA~\cite{kim2024openvla} and the $\pi$ series~\cite{black2024pi_0, intelligence2025pi_}, integrate VLMs into continuous action spaces~\cite{fang2026towards}. Inspired by these advancements, ORION~\cite{fu2025orion} bridges the language between action spaces via a generative planner for continuous trajectories, improving reasoning and generalization in autonomous driving scenarios. DrivePI~\cite{liu2025drivepispatialaware4dmllm} and DriveMonkey~\cite{zhao2025extending} use VLMs that serves as a unified VLA framework to jointly perform spatial understanding, 3D perception, prediction, and planning for AD. However, imitation learning is still limited in effectively propagating trajectory-level supervision into the high-level reasoning space of language models. MindDrive advances this line by integrating online reinforcement learning for adaptive interaction and robust decision-making.

\begin{figure}[t!]
    \centering
    \includegraphics[width=0.98\textwidth]
    {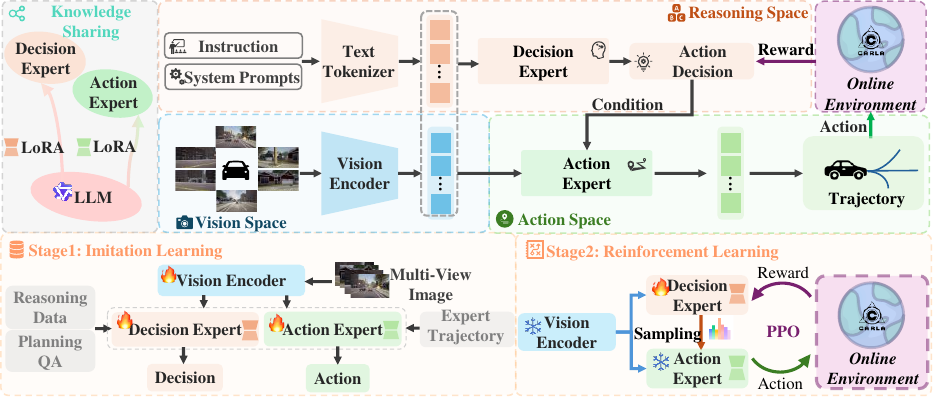}
    \caption{Overview of the MindDrive framework. The system features two experts (Decision and Action) that share a base LLM but employ distinct LoRA adapters. The Decision Expert infers meta-actions from scene, while the Action Expert maps these actions to concrete trajectories. MindDrive is initialized via imitation learning and refined by online closed-loop reinforcement learning to enhance reasoning capabilities.}
    \label{fig: pipeline}
\end{figure}
\subsection{Reinforcement Learning for AD}


Reinforcement learning (RL) has achieved remarkable success across diverse domains, ranging from large language models (LLMs)~\cite{guo2025deepseek} to complex games~\cite{li2025reinforcementgame}, and has recently been increasingly explored in AD~\cite{luo2025adathinkdrive}. CarPlanner~\cite{zhang2025carplanner} leverages expert-guided reward to stabilize training, while RAD~\cite{gao2025rad} and R2SE~\cite{liu2025reinforced} enhance closed-loop training and mitigate catastrophic forgetting. Raw2Drive~\cite{yang2025raw2drive} further introduces a dual-stream model-based RL framework to align raw and privileged world models. Recent works such as AlphaDrive~\cite{jiang2025alphadrive}, AutoVLA~\cite{zhou2025autovla}, and RecogDrive~\cite{li2025recogdrive} employ GRPO-based reinforcement learning~\cite{guo2025deepseek} to refine planning policies through carefully designed reward mechanisms. However, these methods primarily rely on offline expert datasets and handcrafted rewards, lacking actual interaction with dynamic environment. In contrast, online RL enables adaptive decision-making via direct environment interaction and trial-and-error learning but faces slow convergence and inefficient exploration. To address these challenges, we propose an adaptive online RL framework that improves closed-loop robustness and enhances decision-making quality.

\section{Method}

In this section, we present MindDrive in detail. As illustrated in Fig.~\ref{fig: pipeline}, MindDrive comprises two main components: a Decision Expert and an Action Expert, which share a common Vision Encoder and Text Tokenizer but differ only in their respective LoRA parameters~\cite{hu2021lora}. The Decision Expert performs high-level reasoning from navigation instructions and multi-view visual inputs, generating abstract driving decisions in the form of meta-actions. The Action Expert translates these meta-actions into concrete action trajectories, conditioned on the scene information and instruction. This design enables flexible and interpretable action generation, bridging high-level reasoning with low-level control. 

Our training process includes two stages: 1) Imitation learning (IL) establishes a mapping between language and action space (Sec.~\ref{sec: imitation policy learning}), providing high-quality candidate trajectories for online Reinforcement learning (RL) and effectively reducing its exploration space. 2) Online RL further enhances the model's comprehension ability via action rewards in an online environment (Sec.~\ref{sec: reinforcement policy learning}).

\subsection{Problem Formulation}
\label{sec:problem formulation}
In the task of end-to-end autonomous driving, we aim to generate a diverse trajectory set $A$ and determine an optimal trajectory $a^*$ based on surrounding visual information $V$ and language instructions $L$:
\begin{equation}
\begin{aligned}
A &= \left\{{a_i}\right\}_{i=1}^N \quad, \text{where} \quad a_i \sim \pi_g(a \mid V, L),\\
a^{*} &= \arg\max_{a \in A} \pi_g(a \mid V, L),
\end{aligned}
\end{equation}
where $a_i$ represents the trajectory in the multimodal trajectory set $A$ and ${\pi_g}(a \mid V, L)$ is the trajectory generation policy function. Current methods often select the optimal trajectory based on the score-based selection policy $\pi_c$:
\begin{equation}
a^{*} = \arg\max_{a \in A} \pi_c(a).
\end{equation}

Unlike the score-based selection policy, to fully leverage the potential of the VLA model, we model the selection task as an action decision process and introduce $\pi_d(a|V, L)$ as the selection policy function:
\begin{equation} 
\label{eq:formulat}
\begin{aligned}
a^{*} &= \arg\max_{a \in A} \pi_c(a)\\
    &= \arg\max_{a \in A} \underbrace{\pi_d(a \mid V, L)}_{selection} \cdot \underbrace{\pi_g(a \mid V, L)}_{generation}.
\end{aligned}
\end{equation}

From Eq.~\ref{eq:formulat}, we can clearly identify that the generation of the optimal trajectory depends on two core policy functions: $\pi_d$ and $\pi_g$. Previous methods fail to establish a connection between these two policy spaces in online reinforcement learning. To address this challenge, we establish a mapping relationship from linguistic meta-actions to trajectories, and then leverage the trajectory feedback to optimize the reasoning of $\pi_d$ through online reinforcement learning.

\noindent\textbf{Online RL} enables the model to continuously learn and optimize its policy through dynamic interaction with the environment, which is crucial for enhancing the model's understanding of causal relationships. We model our trajectory decision process tasks as a Markov Decision Process (MDP)~\cite{zhang2025carplanner, yang2025raw2drive} for online reinforcement learning. The MDP is structured as a tuple $\langle S, A, P, R, \gamma \rangle $. The state $s_t \in S $ represents all necessary information for the agent to decide at step $t$. The model selects an action from the action space $A$ according to the policy $ (\pi_d,\pi_g)$. Following an action, the system transitions to a new state $s_{t+1}$ according to dynamics implicitly defined by closed-loop simulation environment. The reward $r_t \in R(s_t)$  is a scalar feedback signal that evaluates the quality of executing action in state $s_t$. Our objective is to learn a decision-making policy $\pi_d$ that maximizes the expected cumulative discounted reward in the collected data $\tau$, guided by the discount factor $\gamma$. This objective is formulated as:
\begin{equation} 
J(\theta) = \mathbb{E}_{\tau} \sim {\pi_d} \left[ R(\tau) \right] = \mathbb{E}_{\tau} \sim {\pi_d} \left[ \sum_{t=0}^{T-1} {\gamma}^{t} r_t \right]. 
\end{equation} 

\subsection{Language-Action Mapping}
\label{sec: imitation policy learning}
To enhance the synergy between our decision-making $\pi_d$ and trajectory generation policy $\pi_g$, we decouple a single LLM into two specialized experts with distinct LoRA parameters. One LLM serves as the Decision Expert, implementing the policy $\pi_d$, while the other acts as the Action Expert, responsible for the policy $\pi_g$. This architecture ensures they operate from a shared foundation of world knowledge while performing their distinct functions. We first leverage IL to create a mapping between the Decision Expert and the Action Expert, establishing a connection between language and action and enhancing exploration efficiency in the subsequent reinforcement learning process.

Inspired by~\cite{renz2025simlingo, xu2025drivegpt4}, we decouple the control into longitudinal and lateral control to improve planning flexibility and design the corresponding meta-action in planning QA. We generate the planning QA pairs using LLMs and refine them through manual filtering to ensure a one-to-one correspondence between language and actions (see the Appendix for details). The model is then trained on both the reasoning data and planning QA to learn the mapping from language to action. The loss function can be expressed as:
\begin{equation}
    \mathcal{L}_{CE} = -\frac{1}{N} \sum_{i=1}^{N} \sum_{t=1}^{T_i} \log P(y_{i,t} | x_i, y_{i, <t}).
\end{equation}
where $x$ denotes the input sequence, $N$ denotes the batch size, and $y$ denotes the target sequence, with $y_{i,<t}$ representing historical tokens preceding time step $t$. Then, we map the meta-actions to the temporal speed trajectory for longitudinal control and the geometric path trajectory for lateral control in Action Expert. Specifically, we leverage the autoregressive nature of the Action Expert to encode visual and language information into a hidden state $h$ and introduce two special tokens, $\texttt{<speed\_waypoints>}$ and $\texttt{<path\_waypoints>}$, to extract the logits from the Action Expert's output:
\begin{equation}
P_\theta(h) = \prod_{k=1}^{K} P_\theta(x_k \mid x_{<k}).   
\end{equation}

Finally, we utilize a Variational Autoencoder (VAE)~\cite{kingma2013auto} with a GRU-based decoder to align our language and action spaces, directly translating visual-language representations into the final action trajectory:
\begin{equation}
p(\bm{z}|h) = N(\mathbf{\mu}_h,\mathbf{\sigma}_h^2), 
\pi_g(a|V, L) =gru(\bm{z}),
\end{equation}
where $\bm{z}$ is gaussian variables in the latent space.

We utilize the common detection loss $\mathcal{L}_{det}$ for auxiliary supervision~\cite{fu2025orion}. The VAE is trained by a Kullback-Leibler divergence loss under the supervision of expert trajectories. We employ L1 loss as the Behavior Cloning (BC) loss for speed and path waypoint regression.
The total loss is:

\begin{equation}
\begin{aligned}
\mathcal{L}_{\pi}^{\text{il}}(\theta) =  \mathcal{L}_{BC} + \mathcal{L}_{\text{CE}} + \mathcal{L}_{\text{vae}}+\mathcal{L}_{\text{det}}.
\end{aligned}
\end{equation}

\subsection{Online RL for Action-Reasoning}
\label{sec: reinforcement policy learning}

\begin{wrapfigure}{r}{0.5\textwidth}
    \centering
    \includegraphics[width=0.49\textwidth]{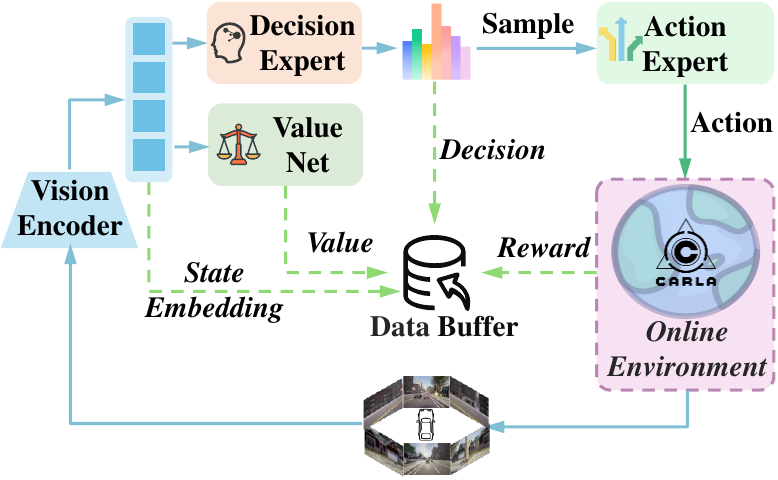}
    \caption{Online reinforcement learning (RL) framework. The RL pipeline stores state embeddings, value estimates, decisions, and action rewards into the data buffer.}
    \label{fig: data collect}
    \vspace{-4pt}
\end{wrapfigure}

IL generates human-like trajectories but often suffers from causal confusion. To overcome this, we leverage online RL within the CARLA simulator~\cite{dosovitskiy2017carla}. As shown in Fig.~\ref{fig: data collect}, this online approach enables the agent to explore the environment through trial and error, learning from direct interaction and consequences, thereby bolstering the model's driving performance in complex scenarios. To leverage prior knowledge from IL, the value network shares the same weights as the LLM, with the only difference being that its final layer is replaced by a Multi-Layer Perceptron (MLP) to predict the state value.

To achieve an efficient rollout process, we deployed $N$ parallel CARLA collectors, focusing on routes across different scenarios that the model failed to complete after IL. At each step, we use a vision encoder to process the scene's visual information and produce state embeddings. We query the Decision Expert with question and sample from its output logits about meta-action tokens. The sampling meta-actions further map to precise trajectories in the action space by the Action Expert. Meanwhile, the value network estimates the value of the current state at each decision step.

Given that MindDrive has already acquired basic driving skills through IL pre-training, we employ a sparse reward function to guide the optimization of its high-level reasoning space. Specifically, a reward of +1 is assigned when the vehicle successfully reaches its destination, while a reward of -1 is incurred when a predefined penalty event is triggered. For all other normal driving scenarios, the reward is 0. The reward function $r_t$ is defined as follows:
\begin{equation}
r_t =
\begin{cases}
  +1 & \text{if the destination is reached} \\
  -1 & \text{if a penalty event is triggered} \\
  0 & \text{otherwise}
\end{cases}.
\end{equation}

We use the official CARLA~\cite{dosovitskiy2017carla} leaderboard metrics as our penalty event. The penalty events include major infractions, such as collisions with other vehicles and running a red light (see more details in Exp.~\ref{Penalty Event}). The rollout process will be terminated once any penalty event is triggered.


After collecting a complete route, the $\delta$ value is calculated by the Temporal-Difference method:
\begin{equation}
\delta_t = r_t + \gamma \cdot V'(s_{t+1}) - V'(s_t),
\end{equation}
where $ V'$ is the value network function, and $\gamma$ is the discount factor to ensure future rewards are appropriately weighted in the decision-making process.
Then we calculate GAE (Generalized Advantage Estimation):
\begin{equation}
    \hat{G}_t = \delta_t + \gamma \lambda  \cdot \hat{G}_{t+1},
\end{equation}
where $\lambda$ is the trace-decay parameter that controls the bias-variance trade-off in advantage estimation. Consequently, the return $\hat{R}_t$ is computed as the sum of the advantage and the value estimate:
\begin{equation}
    \hat{R}_t = \hat{G}_t + V'(s_t).
\end{equation}

Instead of directly using multi-images for RL training, we use the state embedding extracted by the vision encoder to represent the current state. This approach can simultaneously integrate temporal and visual information and improve computational efficiency by avoiding repeated computation. We store the value, decision-action, and reward of each step into a data buffer.

Upon collecting all routes into the rollout buffer, we optimize the policy $\pi_d$ using the Proximal Policy Optimization (PPO) algorithm~\cite{schulman2017proximal}: 
\begin{equation}
\scalebox{0.85}{ 
$ \mathcal{L}_{\text{ppo}} = -\mathbb{E}_t \left[
    \min \left(
        \frac{\pi_d}{\pi_{d_{\text{old}}}} \hat{G}_t,
        \text{clip} \left(\frac{\pi_d}{\pi_{d_{\text{old}}}}, 1 - \epsilon, 1 + \epsilon \right) \hat{G}_t
    \right)
\right], $
}
\end{equation}
where $\epsilon$ is the clipping parameter in PPO. Benefiting from the proposed RL process, MindDrive is able to be trained in a large batch for stable optimization.

Meanwhile, to mitigate the problem of catastrophic forgetting during the RL fine-tuning phase, we introduce a Kullback-Leibler divergence loss as a regularization term. This loss function is designed to constrain the output distribution of the meta-actions of the Decision Expert. Its formula can be expressed as:
\begin{equation}
\mathcal{L}_{KL} = D_{KL}(P_{\text{ref}}(\cdot|s) \parallel P_{\theta}(\cdot|s)). 
\end{equation}

During training, only the parameters of the MLP head within the value net are updated. The optimization is performed by minimizing the Mean Squared Error (MSE) loss, formulated as:
\begin{equation}
\mathcal{L}_{V'}(\theta) = \frac{1}{N} \sum_{t=1}^{N} \left( V'_{\theta}(s_t) - \hat{R}_t \right)^2.
\end{equation}

Finally, the online reinforcement policy learning loss is:
\begin{equation}
\mathcal{L}_{\pi}^{\text{rl}}(\theta) = \mathcal{L}_{ppo}+\mathcal{L}_{V'} + \beta\mathcal{L}_{KL},
\end{equation}
where $\beta$ is a coefficient that controls the strength of the KL regularization.


\section{Experiments}

\subsection{Experimental Settings}
\noindent\textbf{Dataset.} We train and evaluate MindDrive on the Bench2Drive dataset~\cite{jia2024bench2drive}, a comprehensive closed-loop benchmark for autonomous driving (AD) based on the CARLA simulator~\cite{dosovitskiy2017carla}. During imitation learning (IL), we adopt the official base set of 1000 clips, with 950 used for training and 50 for open-loop validation. We perform closed-loop evaluation on the official Bench2Drive benchmark, including 220 short routes across 44 interactive scenarios. For the online reinforcement learning (RL) stage, we utilize a set of 44 rollout routes, where the model successfully reaches the destination during the action sampling phase (see the Appendix for details).

\noindent\textbf{Metrics.}
For closed-loop evaluation, we adopt three key metrics: Driving Score (DS), Success Rate (SR), and Multi-Ability. Following the CARLA benchmark~\cite{dosovitskiy2017carla}, the SR measures the proportion of routes completed within the specified time limit, while DS reflects route completion penalized by traffic infractions. Multi-Ability independently evaluates five advanced urban driving skills. 

\noindent\textbf{Model Setting.} In imitation learning, we set the number of speed waypoints $\mathbf{w_s} \in \mathbb{R}^{N_{s}\times2}$ to $N_s= 6$, representing a future 3-second trajectory sampled at 2Hz. For the path waypoints $\mathbf{w_p} \in \mathbb{R}^{N_{p}\times2}$, $N_p=20$ points are used to define a 20-meter path with a 1-meter interval between consecutive points. The decision space for the Decision VLM includes 7 speed meta-actions and 6 path meta-actions, aligning with the downstream planning trajectory. The detailed definitions of the meta-actions are provided in the Appendix. 


\noindent\textbf{Training Process.} 
All experiments are conducted on 32 NVIDIA A800 GPUs with 80 GB of memory. We employ EVA-02-L~\cite{fang2024eva} as the vision encoder. We use the lightweight Qwen2-0.5B~\cite{qwen2} and Qwen2.5-3B~\cite{qwen2.5} as our base LLM. We use Chat-B2D~\cite{fu2025orion} as our reasoning data and jointly train with planning QA pairs. Both experts are fine-tuned with LoRA~\cite{hu2021lora}, the rank dimension and alpha set to 16. In the online RL stage, data is collected in parallel using 24 Carla simulations. The weights for all penalty components are fixed to unity, and no specialized fine-tuning procedure is adopted. We also propose detailed pseudocode for our online RL algorithm in the Appendix.


\subsection{Main Results}

We conduct a comprehensive comparison between MindDrive and representative methods from both the traditional end-to-end (E2E) and vision-language-action (VLA) paradigms on the Bench2Drive benchmark. Tab.~\ref{tab: main_table} lists the detailed results, and we find that:

\textbf{1) MindDrive delivers promising performance with a lightweight model.} Compared with the traditional E2E methods, MindDrive surpasses the latest state-of-the-art (SOTA) IL model PGS~\cite{huang2025prioritizing} by 6.45\% SR, and the online RL-based method Raw2Drive~\cite{yang2025raw2drive} by 6.68 DS and 4.85\% SR. With a 3B parameter LLM, MindDrive-L outperforms the PGS~\cite{huang2025prioritizing} method by 2.51 in DS and delivers a 9.62\% gain in SR. Within the VLA paradigm, MindDrive achieves comparable performance to the IL-based ORION~\cite{fu2025orion} and outperforms DriveMoE~\cite{yang2025drivemoe} by 3.82 DS and 6.45\% SR. In open-loop metrics, MindDrive exhibits a performance gap compared to DriveMoE. We attribute this to differing optimization objectives which DriveMoE employs a skill-specialized action MoE that explicitly mitigates mode averaging during trajectory fitting to directly optimize open-loop performance. Conversely, MindDrive incorporates online RL rewards designed to prioritize closed-loop driving performance. Notably, using the lightweight Qwen2-0.5B, MindDrive achieves competitive performance in contrast to the substantially larger LLMs employed by ORION and DriveMoE (\textit{i.e.}, Vicuna1.5-7B and Paligemma-3B), highlighting our method's superior efficiency. 


\input{table/main_table}

\textbf{2) Online RL enhances MindDrive’s capabilities for complex, dynamic interactions.} As shown in Tab.~\ref{tab: main_table}, MindDrive demonstrates a clear advantage over other methods. It surpasses the offline RL-based RecogDrive~\cite{li2025recogdrive} by 6.68 DS and 9.64\% SR, and MindDrive-L outperforms the SOTA RL-based method AutoVLA~\cite{zhou2025autovla} by 1.75 DS and 0.53\% SR. Furthermore, MindDrive outperforms the IL-based MindDrive 2.19 DS and 5.79\% SR, while MindDrive-L surpasses the IL-based MindDrive-L 4.81 DS and 7.55\% SR, strongly validating the superiority of the proposed online RL paradigm. 

The Multi Ability metric further supports this finding. Specifically, MindDrive achieves a 14.91\% improvement in mean ability over RecogDrive~\cite{li2025recogdrive} and 5.57\% over the IL method ORION~\cite{fu2025orion} with the same lightweight LLM. Notably, it achieves substantial gains in capabilities closely tied to meta-actions selection, with improvements of 55.56\% in Overtaking and 30\% in Give Way compared to the offline RL method RecogDrive. Although MindDrive performs lower than the SOTA VLA methods in the Emergency Brake and Traffic Sign ability, it still exhibits a noticeable advantage over our IL-based version, with improvements of 8.33\% and 0.98\%, respectively. Furthermore, MindDrive-L achieves new SOTA performance on the DS, SR, and mean driving ability metrics. These results confirm that online RL significantly strengthens the model's causal reasoning and decision-making robustness in complex interactive environments.

\subsection{Ablation Studies}
In ablation studies, we use 0.5B parameter LLM model. For each route, two online reinforcement learning epochs are performed unless otherwise specified.

\input{table/ab_events}

\noindent\textbf{Ablation on Penalty Events.}
\label{Penalty Event}
We introduce four penalty events during the online RL stage: collisions with pedestrians or vehicles, running a red light, driving off-road, or deviating over 30 meters from the route, and failing to obey a stop sign (\textit{i.e.}, Collision, Traffic Light, Route Deviation, and Stop, respectively). We assign a sparse reward of -1 to penalize our model for triggering these events. As shown in Tab.~\ref{tab: ab_events}, the progressive incorporation of these penalties leads to consistent improvements in both the success rate and the mean driving ability score relative to the IL baseline (ID-1).

Specifically, introducing the collision penalty (ID-2) yields a 1.4\% SR and 3.76\% mean ability improvement over the IL baseline (ID-1), while maintaining a comparable DS. In overtaking scenarios, MindDrive demonstrates exceptional performance, outperforming the baseline (ID-1) by a significant margin of 4.44\%. We attribute this improvement to the model's learned ability to adopt a more proactive strategy for collision avoidance in the continuous and interactive traffic flow. However, this strategic shift consequently impairs its merging performance.

Incorporating the traffic-light penalty (ID-3) brings further improvements over the IL baseline (ID-1), elevating Traffic Sign by 1.52\% and Emergency Braking by 8.97\%. However, conflicting reward signals within this penalty lead to a noticeable drop in overtaking performance. Introducing a route-deviation penalty (ID-4) helps achieve a better trade-off between decisiveness and caution, although the stricter constraints on exploration limit further performance gains. Notably, adding a stop sign penalty significantly boosts overall performance, as it strongly correlates with the stop meta-action, enabling more effective policy learning. This is especially beneficial in merging scenarios that feature stop signs, yielding improvements of 5.26\% in Merging ability and 3.24\% in the SR compared to ID-4.

Without complex reward engineering, MindDrive can discover effective driving policies through online trial-and-error, autonomously learning from failures to progressively determine optimal actions. We argue that by building upon the foundational driving skills acquired via IL, MindDrive operates effectively in a binary outcome setting where an episode either succeeds or fails. In this setup, the sparse reward from the final outcome is directly propagated back to each decision step, providing sufficient supervision to distinguish between optimal and suboptimal actions. Furthermore, our replay buffer retains both successful and failed episodes, and the large batch size amplifies the contrast between trajectories of different outcomes and lengths, providing sufficient gradient signals for policy improvement.

\begin{table}[tp!]
\centering
\caption{Ablation studies on policy regularization and control methods. DS and SR denote Driving Score and Success Rate, respectively.}
\label{tab: ablation_studies}
\footnotesize

\begin{minipage}[t]{0.48\textwidth}
    \centering
    (a) Ablation on policy regularization \label{tab: effectiveness of different policy}\\[1.5mm] 
    \begin{tabular}{c c c}
    \toprule
    Policy Regularization & DS $\uparrow$  & SR(\%) $\uparrow$ \\
    \midrule
    PPO-Vanilla & 74.73 & 46.73 \\
    PPO-Entropy & 75.71 & 49.24 \\
    PPO-KL      & \textbf{78.04} & \textbf{55.09} \\ 
    \bottomrule
    \end{tabular}
\end{minipage}
\hfill
\begin{minipage}[t]{0.50\textwidth}
    \centering
    (b) Ablation on control methods \label{tab: meta-action}\\[1.5mm]
    \begin{tabular}{c c c}
    \toprule
    Control Method & DS $\uparrow$  & SR(\%) $\uparrow$ \\
    \midrule
    Navigation Command (IL) & 68.11 & 41.59 \\
    Meta Action (IL)        & 75.85 & 49.30 \\ 
    Meta Action (RL)        & \textbf{78.04} & \textbf{55.09} \\ 
    \bottomrule
    \end{tabular}
\end{minipage}
\end{table}

\noindent\textbf{Ablation on Policy Regularization.} We evaluate different policy regularization methods within PPO, as reported in Tab.~\ref{tab: effectiveness of different policy} (a). Our method outperforms vanilla PPO by 3.31 DS and 8.36\% SR, demonstrating that KL divergence loss effectively stabilizes policy updates during RL training and mitigates catastrophic forgetting. Compared to entropy-based regularization (PPO-Entropy), our approach improves by 2.33 DS and 5.85\% SR, indicating that while entropy regularization promotes exploration, excessive policy stochasticity is suboptimal for goal-directed driving tasks. Overall, our KL regularization method enables more efficient learning, with a faster policy optimization process and higher sample efficiency than the baselines.

\noindent\textbf{Ablation on Control Method.} We further investigate the effects of different control methods by comparing two high-level instruction approaches: navigation commands and LLM-generated meta-actions. As detailed in Tab.~\ref{tab: meta-action} (b), the VLM-guided meta-actions IL model achieves 7.74 DS and 7.71\% SR improvements over the navigation-command baseline, indicating that VLM-derived meta-actions enable more effective reasoning in complex traffic scenarios. Meanwhile, employing online RL further improves meta-action selection, yielding additional gains of 2.19 DS and 5.79\% SR, which demonstrates that real-time interactive reinforcement learning can dynamically optimize the reasoning logic of vision-language meta-actions and better adapt to the constantly changing traffic environments.

\begin{wrapfigure}{r}{0.5\textwidth}
    \centering
    \includegraphics[width=0.48\textwidth]{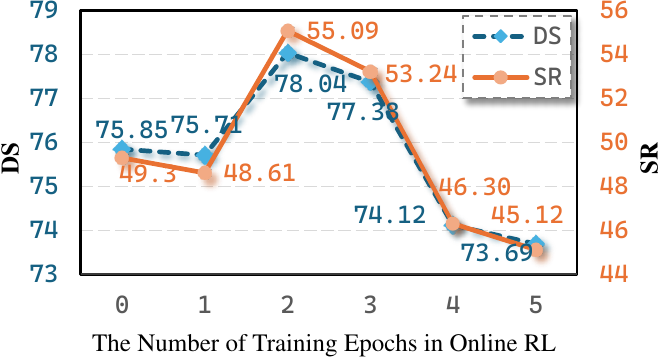}
    \caption{Ablation on the Training Epochs for Online RL. DS and SR denote Driving Score and Success Rate, respectively.}
    \label{fig: route_number}
    \vspace{-2pt}
\end{wrapfigure}

\noindent\textbf{Ablation on the Number of Training Epochs.} We finally analyze MindDrive's sensitivity to the number of training epochs during online RL. As shown in Fig.~\ref{fig: route_number}, when only one epoch is trained, the value network's inaccurate estimation leads to wrong advantage estimates for each action, resulting in a performance drop compared to the baseline. After two training epochs, our model significantly outperforms the baseline, improving DS by 2.19 and SR by 5.79\%.  However, further increasing the epochs substantially degrades performance, with DS dropping from 78.04 to 73.69 and SR declining from 55.09\% to 45.12\%. We attribute this degradation to catastrophic forgetting, where excessive training causes the policy to overfit recent experiences and forget previously learned scene understanding (See the Appendix for full analysis).
MindDrive further enhances the model's driving capabilities via online RL, achieving optimal performance in just two online training epochs, thereby demonstrating the high efficiency of our online RL framework. Therefore, we set the default number of training epochs to 2, balancing sample efficiency and training stability.

\begin{figure}[ht]
    \centering
    \includegraphics[width=0.98\textwidth]{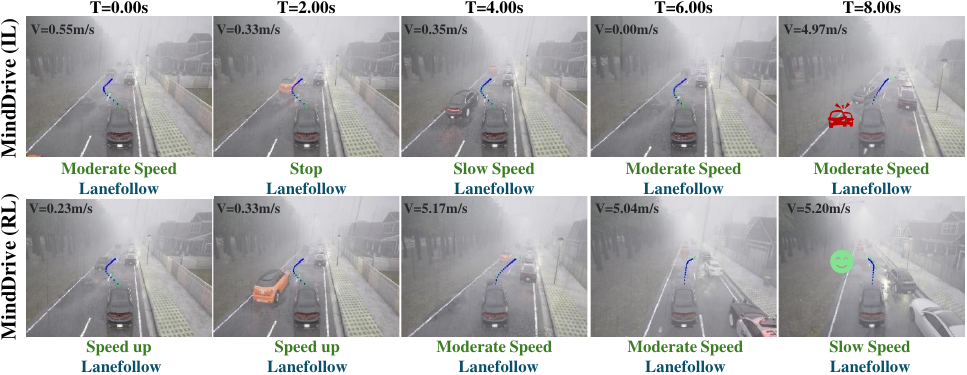}
    \caption{Qualitative results of MindDrive after IL and RL training on the closed-loop evaluation of the Bench2Drive lanefollow scenario. The \textcolor{green}{green} and \textcolor{blue}{blue} refer to the prediction speed and path meta-actions, respectively. \textcolor{red}{Red} denote a collision has occurred.}
    \label{fig: visulation}
\end{figure}

\subsection{Qualitative Results}
\label{qualitative results}

We present a qualitative comparison between the IL and online RL versions of MindDrive in the Bench2Drive lane-follow scenario, as shown in Fig.~\ref{fig: visulation}. The IL paradigm exhibits strong task-specific competence, such as issuing timely stop commands for early braking. However, IL-only MindDrive struggles in dynamic and interactive scenarios, especially in those requiring complex decision-making (\textit{e.g.}, determining the optimal timing for lane changes). After Online RL training, MindDrive selects more robust meta-actions in challenging scenarios, resulting in safer and more decisive lane-change behaviors. 

These qualitative results demonstrate that the online RL stage substantially enhances the VLM’s high-level reasoning and decision-making capabilities, enabling it to better handle complex and uncertain traffic environments.

\section{Conclusion}
In this paper, we introduce \textbf{MindDrive}, a novel autonomous driving framework that leverages language as an interface for online reinforcement learning (RL). MindDrive maps language instructions to actions, transforming the exploration space into a discrete language space to reduce the cost of online RL. It further enables the Large Language Model to refine its reasoning through action feedback in a closed-loop simulator. Extensive experiments within our proposed online RL training framework demonstrate that MindDrive achieves state-of-the-art performance with a lightweight model. To the best of our knowledge, this is the first work to successfully train a Vision-Language-Action model for autonomous driving in an interactive simulator. We expect this work will provide valuable insights for the autonomous driving community.


\noindent\textbf{Limitations.}
Due to the absence of real-world interactive simulators for autonomous driving, our evaluation is currently limited to the CARLA simulator~\cite{dosovitskiy2017carla}. Additionally, the challenge of synchronizing multiple CARLA simulators prevents us from evaluating multiple candidate actions from an identical initial state across parallel environments, restricing the application of the GRPO~\cite{shao2024deepseekmath} algorithm in our framework. Future work will explore real-world interactive simulation infrastructures or alternative training strategies to better support parallel policy evaluation and more efficient online reinforcement learning.

\noindent\textbf{Acknowledgment.} 
This work is supported by the NSFC (62225603, 623B2038).
%
%
\bibliographystyle{splncs04}
\bibliography{main}

\clearpage
\setcounter{page}{1}
\appendix

\section*{Supplementary Material of MindDrive}

\section{Overview}
The supplementary material is organized as follows:
\begin{itemize}
    \item Appendix~\ref{sup:question}: Potential Questions Raised by the Main Text.
    \item Appendix~\ref{sup: Planning QA}: Detailed Information on Planning QA Generation.
    \item Appendix~\ref{sup: training_details}: Online Reinforcement Learning Training Details.
    \item Appendix~\ref{sup: more_result}: Additional Experimental Results.
\end{itemize}

\section{Questions}
\label{sup:question}
\noindent\textbf{Q1.} \textit{What are the differences between the framework proposed in this paper and Orion?}

Orion focuses on the holistic optimization of semantic understanding and action execution within a unified framework. In contrast, our work concentrates on effectively applying online Reinforcement Learning (RL) to Vision-Language-Action (VLA) models. While Orion employs a VLM and a generative planner to produce trajectories, our architecture utilizes a dual-expert system that enhances the VLA's reasoning capabilities through online action feedback and environmental interaction. 

\noindent\textbf{Q2.} \textit{Why does the method in the paper claim to be exploration-efficient?}

Autonomous driving requires selecting optimal control signals from an infinite, continuous action space, which poses a significant challenge for exploration. By mapping the infinitely continuous action space to a discrete language space of meta-actions, we significantly reduce the complexity of the learning problem. Given that online RL faces severe efficiency bottlenecks in autonomous driving, this discretization is a critical design choice that enables us to practically deploy and evaluate VLA-based methods in this domain.

\noindent\textbf{Q3.} \textit{Is it necessary to employ two experts?}

In the single-expert setup, jointly optimizing semantic understanding and precise action generation often causes catastrophic forgetting~\cite{zhou2025chatvla}. The $\pi$ series~\cite{intelligence2025pi,intelligence2025pi_} mitigate this by leveraging knowledge insulation and supervision from human demonstrations during their RL process. However, online RL in autonomous driving tasks lacks correct trajectory supervision. Consequently, the process of updating VLA reasoning interferes with trajectory generation, leading to a catastrophic collapse in closed-loop performance (see Sec.~\ref{Single Expert} ). To address this, we adopt a dual-expert architecture with two LoRA modules, updating only minimal parameters (2M) and incurring limited inference latency (50 ms).

\noindent\textbf{Q4.} \textit{How to ensure alignment between discrete meta-actions and trajectories?}

To ensure rigorous alignment between discrete meta-actions and actual planned trajectories, we first verify during data annotation that each QA sample maintains a perfect correspondence between the labeled meta-action and the ground-truth trajectory. This deterministic one-to-one mapping is then solidified during the Imitation Learning phase, where the Decision Expert learns to output meta-actions and the Action Expert learns their corresponding trajectories. Consequently, during the Reinforcement Learning phase, this learned mapping remains fixed, ensuring that any meta-action sampled by the Decision Expert is instantly and accurately executed as a trajectory in the CARLA simulator.

\noindent\textbf{Q5.} \textit{What is the difference between Online and Offline RL for Autonomous Driving?}

We compare the online reinforcement learning (RL) approach with offline RL\begin{wrapfigure}{r}{0.5\textwidth}
    \vspace{-20pt}
    \centering
    \includegraphics[width=0.49\textwidth]{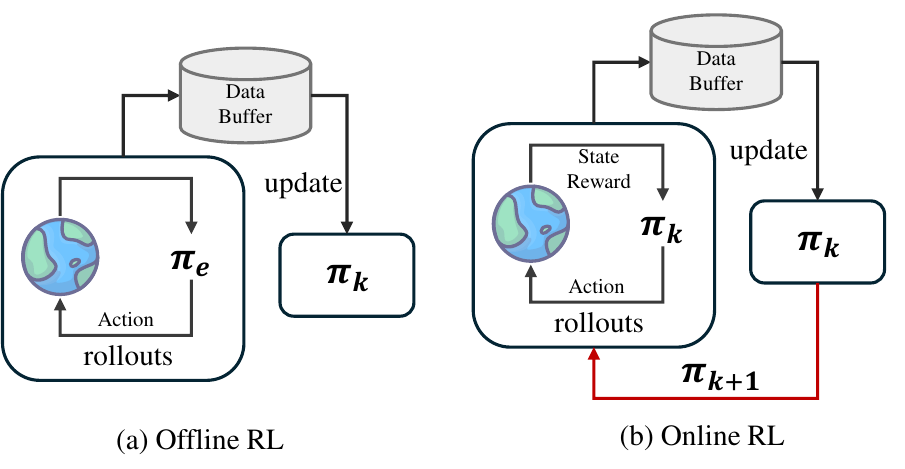}
    \caption{A Comparison of Online and Offline Reinforcement Learning (RL). $\pi_e$ and $\pi_k$ represent distinct policies.}
    \label{ofon}
    \vspace{-15pt}
\end{wrapfigure} in autonomous driving, as illustrated in Fig.~\ref{ofon}. Offline RL uses a fixed dataset collected by an expert policy $\pi_e$. The updated policy model $\pi_k$ cannot be deployed into the online environment to gather new data through interaction. This paradigm often leads to causal confusion and distributional shift. In contrast, online RL deploys the updated policy $\pi_{k+1}$ back into the online environment for interaction. Through trial and error in the dynamic environment, it collects new data to further strengthen the policy.

\section{The Planning QA Generation Process}
\label{sup: Planning QA}

\begin{wrapfigure}{r}{0.5\textwidth}
    \vspace{-15pt}
    \centering
    \includegraphics[width=0.49\textwidth]{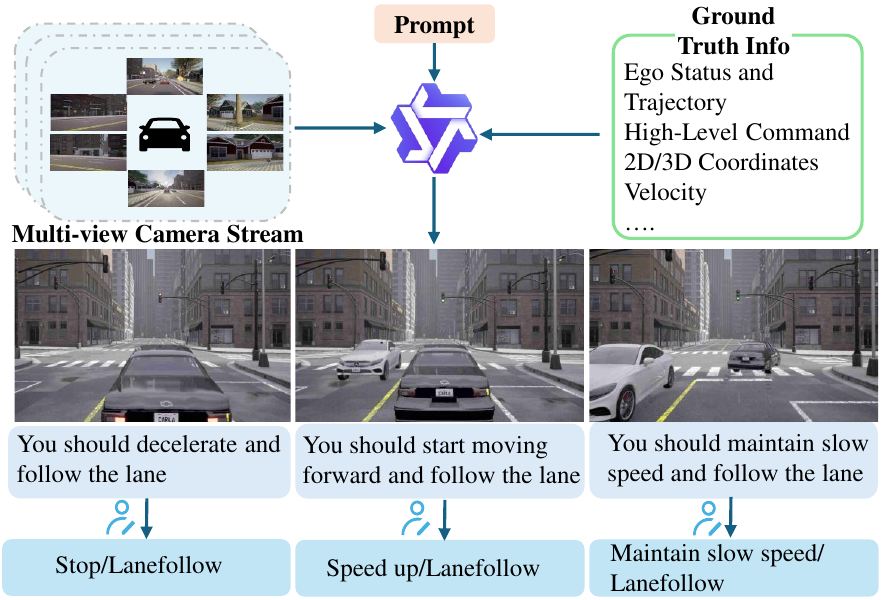}
    \caption{The production process of meta-actions. Based on consecutive frames, prompt, and the current frame's ground truth, the VLM generates a scene action judgment, which is subsequently manually translated into meta-actions.}
    \label{fig_data_process}
    \vspace{-20pt}
\end{wrapfigure}
To construct the planning QA, we decompose the ego vehicle's control actions into seven speed-control meta-actions and six path-control meta-actions. This decomposition enables more precise and flexible control in complex traffic scenarios:
\begin{itemize}
    \item \textbf{Speed Action}: Speed Up; Slow Down; Slowdown Rapidly; Maintain Slow Speed; Maintain Moderate Speed; Maintain Fast Speed; Stop.
    \item \textbf{Path Action}: Turn Left; Turn Right; Change Lane Left; Change Lane Right; Straight; Lanefollow.
\end{itemize}
To obtain the meta-actions for each frame, we implement a two-stage data annotation pipeline, as illustrated in Fig.~\ref{fig_data_process}. First, we employ Qwen2VL-72B~\cite{wang2024qwen2vl} to generate decision-making conditioned on the current scenario, prompt, and ground-truth information. Second, the VLM's output action semantics undergo rule-based filtering to ensure strict alignment with standardized meta-actions and ground truth trajectory, balancing comprehensiveness and precision. 

\begin{figure}[t!]
    \centering
    \includegraphics[width=0.90\textwidth]{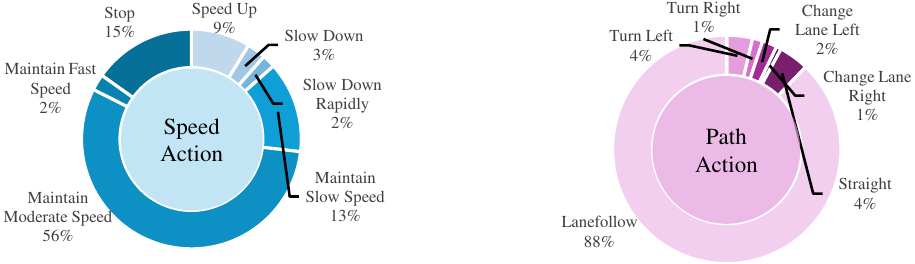}
    \caption{The statistical characteristics of meta-actions}
    \label{data_statistics1}
    \vspace{-16pt}
\end{figure}

Specifically, for speed, we first determine the appropriate meta-action based on the current speed $v_{\text{cur}}$, as shown in the Eq.~\ref{eq:v}. 

\begin{equation}
    S =
\begin{cases}
\text{stop}, & |v_{\text{cur}}| < 0.05 \\
\text{slow}, & 0.05 \leq v_{\text{cur}} < 5 \\
\text{moderate}, & 5 \leq v_{\text{cur}} < 10 \\
\text{fast}, & v_{\text{cur}} \geq 10
\end{cases}
\label{eq:v}
\end{equation}

The acceleration state \( A \) is derived from the speed variation \( \Delta v = v_i - v_0 \) (where \( v_i \) denotes the speed of the \( i \)-th subsequent frame and \( v_0 \) is the initial speed of the current frame) and the relative speed change rate \( r = \frac{\Delta v}{\max(|v_0|, 1)} \).

\begin{equation}
A =
\begin{cases}
\text{speed up}, & \Delta v > 0.2 \land r > 0.15 \\
\text{slow down rapidly}, & \Delta v < -1.0 \land r > 0.8 \\
\text{slow down}, & \Delta v < -0.3 \land r > 0.4 \\
\text{maintain } S \text{ speed}, & \text{otherwise}
\end{cases}
\end{equation}

Speed actions are determined by combining the current speed and the acceleration trend, yielding a discrete and semantically consistent speed action for planning QA. For path action, we directly use the Ground Truth (GT) command from Bench2Drive.

We also perform a statistical analysis on the generated meta-actions, as shown in Fig.~\ref{data_statistics1}. We collect a total of 234,769 planning QA samples. The data distribution truly reflects the statistical patterns of driving behavior in the Bench2Drive dataset~\cite{jia2024bench2drive} and comprehensively covers various actions required for safe driving.

\begin{table}[t!]
    \centering
    \tiny
    \caption{The collected driving scenarios for reinforcement learning rollout}
    \label{tab:RL routes}
    \setlength{\tabcolsep}{0.1mm}
    \renewcommand{\arraystretch}{0.8} 
    \begin{tabular}{c c c c c c c c c}
        \toprule
        Route\_id & Scenario & Skill & Route\_id & Scenario & Skill & Route\_id & Scenario & Skill \\
        \midrule
        14194 & {\makecell{Pedestrian\\Crossing}} & \makecell{Traffic\\Sign} & 
        24071 & {\makecell{Interurban\\Advanced\\ActorFlow}} & Merging & 
        28210 & \makecell{NonSignalized\\JunctionLeftTurn\\EnterFlow} & \makecell{Traffic\\Sign} \\
        \midrule
        17563 & \makecell{Sequential\\LaneChange} & Merging & 
        24098 & \makecell{Interurban\\ActorFlow} & Merging & 
        2898 & \makecell{Blocked\\Intersection} & Merging \\
        \midrule
        17569 & \makecell{Sequential\\LaneChange} & Merging & 
        25300 & \makecell{HazardAt\\SideLane} & Overtaking & 
        3178 & \makecell{VanillaNon\\SignalizedTurn} & \makecell{Emergency\\Brake}\\
        \midrule
        17635 & \makecell{Sequential\\LaneChange} & Merging & 
        3307 & Accident & \makecell{Traffic\\Sign} & 
        2164 & \makecell{VehicleTurning\\RoutePedestrian} & \makecell{Emergency\\Brake} \\
        \midrule
        17773 & \makecell{Parked\\Obstacle} & Overtaking & 
        25439 & \makecell{HazardAt\\SideLane} & Overtaking & 
        3380 & \makecell{YieldTo\\EmergVehicle} & Overtaking\\
        \midrule
        1792 & \makecell{HazardAt\\SideLane} & Overtaking & 
        2554 & \makecell{Parked\\Obstacle} & Overtaking & 
        3410 & \makecell{Accident\\TwoWays} & \makecell{Give\\Way}\\
        \midrule
        1656 & \makecell{Parking\\Exit} & Merging & 
        25854 & \makecell{HazardAtSide\\LaneTwoWay} & Overtaking & 
        3514 & \makecell{Parking\\Exit} & Overtaking \\
        \midrule
         2534 & Accident & Overtaking & 
        25863 & \makecell{Pedestrian\\Crossing} & Overtaking & 
        3714 & \makecell{VehicleTurning\\RoutePedestrian} & Merging\\
        \midrule
        2201 & \makecell{Enter\\ActorFlow} & \makecell{Traffic\\Sign} & 
        2606 & \makecell{Construction\\ObstacleTwoWays} & \makecell{Emergency\\Brake} & 
        3785 & \makecell{MergeInto\\SloTraffic} & \makecell{Emergency\\Brake}\\
        \midrule
        23670 & \makecell{Highway\\Exit} & Merging & 
        26950 & \makecell{OppositeVehicle\\RunningRedLight} & Overtaking & 
        3876 & \makecell{VanillaSignalized\\TurnEncounter\\RedLight} & Merging\\
        \midrule
        28085 & T\_junction & \makecell{Emergency\\Brake} & 27018 & \makecell{Signalized\\Junction\\RightTurn} & \makecell{Emergency\\Brake} & 
        3904 & \makecell{VanillaNonSignalized\\TurnEncounter\\Stopsign} & \makecell{Traffic\\Sign}\\
        \midrule
        23901 & \makecell{Interurban\\ActorFlow} & Merging & 
        27529 & \makecell{Pedestrian\\Crossing} & \makecell{Traffic\\Sign} & 
        3936 & \makecell{Signalized\\Junction\\LeftTurn} & \makecell{Traffic\\Sign} \\
        \midrule
        23930 & \makecell{Interurban\\ActorFlow} & Merging  & 
        2373 & \makecell{VanillaSignalized\\TurnEncounter\\RedLight} & \makecell{Traffic\\Sign} &
        4468 & \makecell{Signalized\\Junction\\LeftTurn} & Merging\\
        \midrule
        24041 & \makecell{Highway\\Exit} & Merging & 
        28049 & \makecell{SignalizedJunction\\LeftTurn\\EnterFlow} & \makecell{Traffic\\Sign} & 
        4683 & \makecell{Signalized\\Junction\\LeftTurn} & Merging\\
        \midrule
        28098 & T\_junction & \makecell{Traffic\\Sign} & 2397 & \makecell{VanillaNon\\SignalizedTurn} & \makecell{Traffic\\Sign}\\
        
        \bottomrule
    \end{tabular}
    \vspace{-10pt}
\end{table}


\begin{table}[t!]
  \centering
\caption{Comparison of skill distribution with Bench2Drive}
\vspace{-8pt}
\small
  \label{tab:skill_dis}
  \setlength{\tabcolsep}{1.8mm}
  \renewcommand{\arraystretch}{0.9}
    \begin{tabular}{c c c c c c}
      \toprule
      Scenarios &Merging & Overtaking & Emergency Brake & Give Way & Traffic Sign \\
      \midrule
      Bench2Drive &28.1\% & 15.8\%  & 21.1\% & 3.5\% & 31.6\% \\
      Ours &36.4\% & 22.7\%  & 13.6\% & 2.3\% & 25.0\% \\
      \bottomrule
    \end{tabular}
    \vspace{-14pt}
\end{table}

\section{Training Details}
\label{sup: training_details}
We select 44 routes that the model successfully completes through sampled actions, and each routes rollout five times. The selected route IDs, scenario type and corresponding skills are shown in Tab.~\ref{tab:RL routes}. We also count the skills included in the scenarios of our 44 rollout routes and compare their coverage with that of the test routes in Bench2Drive. As reported in the Tab.~\ref{tab:skill_dis}, the results verify\begin{wraptable}{r}{0.4\textwidth}
    \vspace{-12pt}
    \centering
    \caption{Hyperparameters of model architecture, PPO-related parameters, and loss weights.}
    \label{table:ppo_param}
    \small 
    \setlength{\tabcolsep}{2.0mm}
    \renewcommand{\arraystretch}{1.0} 
    \begin{tabular}{l c}
    \toprule
    {Parameter} & Value \\
    \midrule
    discount factor $\gamma$ & 0.99 \\
    GAE parameter $\lambda$ & 1.0 \\
    Clip range $\epsilon$ & 0.2 \\
    Batch size $B$ & 32 \\
    \midrule
    Weight of value loss & 0.5 \\
    Weight of KL loss & 0.5 \\
    Weight of PPO loss & 1.0 \\
    \bottomrule
    \end{tabular}
    \vspace{-16pt}
\end{wraptable} that these 44 rollout routes have sufficient skill coverage.

To efficiently collect diverse and representative trajectory data, we implement a parallelized data collection framework using 24 instances of the CARLA simulator. Each simulator instance executes independent driving episodes with varied initial conditions. The episodes are terminated either upon successful task completion or when a predefined penalty condition is triggered. The data collection process takes approximately 24 hours.

Subsequently, We train the policy using large batches and advantage regularization to update the policy and value networks within the Proximal Policy Optimization (PPO)~\cite{schulman2017proximal} algorithm. The key hyperparameters, including batch size, discount factor, Generalized Advantage Estimation (GAE) lambda, and clip range, are detailed in Tab.~\ref{table:ppo_param}. The pseudocode for our online reinforcement learning is presented in the Algorithm~\ref{algorithmic}.

\begin{algorithm}[t!]
\caption{Online Reinforcement Learning Procedure of MindDrive.} 
\label{algorithmic}
\hrule
\begin{algorithmic}[1]
\State \textbf{Input:} RL epochs $E_\text{rl}$, parallel environments $N_\text{env}$
\State \textbf{Require:} Vision Encoder $\Phi$, Base LLM $\Theta$, Action Expert $\pi_g$, Decision Expert $\pi_d$, Value Network $V'$, Reference Policy $\pi_{\text{ref}}$

\For{$\text{epoch} = 1$ \textbf{to} $E_\text{rl}$}
    \State Initialize rollout buffer $\mathcal{B} \leftarrow \emptyset$
    
    \State \textcolor{gray}{\# Phase 1: Trajectory Rollout}
    \For{$\text{each env}$ \textbf{in} $N_\text{env}$}
        \State Reset env to get initial visual observation $I_0$, and encode state $s_0 = \Phi(I_0)$
        \While{episode not done}
            \State Sample meta-action $c_t \sim \pi_d(\cdot| s_t)$, get value $V'_t = V'(s_t)$
            \State Execute trajectory $\tau_t = \pi_g(\cdot| s_t, c_t)$
            \State Observe reward $r_t$ and next visual input $I_{t+1}$, encode $s_{t+1} = \Phi(I_{t+1})$
            \State Store transition $(s_t, c_t, r_t, V'_t, \pi_d(c_t|s_t))$ in $\mathcal{B}$
        \EndWhile
    \EndFor
    
    \State \textcolor{gray}{\# Phase 2: Advantage Estimation}
    \State Compute GAE advantages $\hat{G}_t$ and value targets $\hat{R}_t$ for all transitions in $\mathcal{B}$
    
    \State \textcolor{gray}{\# Phase 3: PPO Policy Optimization}
    \For{mini-batch \textbf{in} $\mathcal{B}$}
        \State Compute PPO loss $\mathcal{L}_{\text{clip}}$, KL penalty $\mathcal{L}_{\text{KL}}$, and value loss $\mathcal{L}_{\text{value}}$ via Eq.13 - Eq.15
        \State Compute total RL objective $\mathcal{L}_{\text{rl}}$
        \State Update trainable parameters ($\pi_d$ and $V'$) via backpropagation on $\mathcal{L}_{\text{rl}}$
    \EndFor
\EndFor

\State \textbf{Output:} RL-optimized Decision Expert $\pi_d$ and Value Network $V'$
\end{algorithmic}
\hrule
\end{algorithm}

\section{More Results}
\label{sup: more_result}
\subsection{Runtime Analysis}
We performed a quantitative runtime analysis of MindDrive on an NVIDIA A100 GPU and report the results in Tab.~\ref{tab:runtime_per_component}. Under the lightweight 0.5B setting, the model achieves a average inference time of approximately 540 ms, where the Vision Encoder accounts for about 400 ms and the two language experts each contribute about 50 ms. Compared with prior VLA baselines, this latency is substantially lower than Orion ($\sim$1106 ms) and is also faster than RecogDrive ($\sim$750 ms). These results indicate that MindDrive provides a favorable efficiency-performance trade-off, maintaining competitive driving quality with relatively low inference overhead.
\begin{table}[t!]
    \centering
    \caption{Comparison of Model Running Time and Performance. The runtime breakdown of our MindDrive is detailed in the bottom section. Best results are in \textbf{bold}.}
    \label{tab:runtime_per_component}
    \small
    \setlength{\tabcolsep}{2.8mm}
    \renewcommand{\arraystretch}{1.0}
    \begin{tabular}{l c c c}
        \toprule
        Model & Mean Runtime (ms) & DS (\%) & SR (\%) \\
        \midrule
        Orion (ICCV25)      & $\sim 1106$ & 77.74 & 54.62 \\
        RecogDrive (ICLR26) & $\sim 750$  & 71.36 & 45.45 \\
        \textbf{MindDrive (Ours)} & $\mathbf{\sim 540}$ & \textbf{78.04} & \textbf{55.09} \\
        
        \midrule
        \midrule
        \multicolumn{4}{c}{\textit{Runtime Breakdown of MindDrive}} \\
        \midrule
        {Component} & {Mean Runtime (ms)} & \multicolumn{2}{c}{{--}} \\
        \midrule
        {Vision Encoder} & {$\sim 400$} & {--} & {--} \\
        {Decision Expert (Qwen2-0.5B)} & {$\sim 50$} & {--} & {--} \\
        {Action Expert (Qwen2-0.5B)} & {$\sim 50$} & {--} & {--} \\
        \bottomrule
    \end{tabular}
\end{table}   

\subsection{Experiments on Rollout Routes}
\label{Rollout Path}
\begin{wraptable}{r}{0.5\textwidth}
\vspace{-33pt}
\centering
\caption{Comparative analysis of performance on rollout routes and remain routes. DS and SR denote Driving Score and Success Rate, respectively.}
\label{tab:Rollout}
\footnotesize
\setlength\tabcolsep{2.8mm} 
\begin{tabular}{ c   c c }
\toprule
\multirow{2.6}{*}{\shortstack{Model}}  & \multicolumn{2}{c}{\textbf{Closed-loop}}  \\ 
 \cmidrule(lr){2-3}  & DS$\uparrow$ & SR(\%)$\uparrow$  \\ 
\midrule
\multicolumn{3}{c}{\textit{Rollout Routes}} \\
\midrule
MindDrive (IL) & 75.40 & 29.55  \\
MindDrive (RL) & 80.69 & 38.64  \\
\midrule
\multicolumn{3}{c}{\textit{Other Routes}} \\
\midrule
MindDrive (IL) & 76.26 & 54.71  \\
MindDrive (RL) & 77.33 & 59.41  \\
\bottomrule
\end{tabular}
\vspace{-10pt}
\end{wraptable}
To evaluate the effectiveness of reinforcement learning, we analyze the model’s driving performance over both the rollout routes and the remaining routes. As shown in Tab.~\ref{tab:Rollout}, after reinforcement learning, the model not only achieves higher success rates on the rollout route set but also shows improvements on the other paths. Specifically, along the rollout routes, our model outperforms imitation learning by 5.29 in driving score and 9.09\% in driving success rate, demonstrating its ability to make superior driving decisions by trial-and-error experience. On the remaining routes, our model still achieves improvements of 1.06 in driving score and 4.7\% in success rate over imitation learning, indicating that our reinforcement-based approach possesses strong generalization capabilities beyond the training distribution.
\input{table/supp/sup_route_compare}

Meanwhile, we also conduct a detailed comparison between the model after Online RL and IL on the rollout routes. As detailed in Tab.~\ref{tab:route_performance}, Online RL outperforms IL in most driving scenarios, particularly excelling in overtaking and emergency braking situations. Notably, it elevates the success rate to 100\% in routes 17773, 25863, 3307, 3514, and 3714. Despite this overall improvement, performance degradation is observed in some specific paths after applying Online RL. We argue that this occurs because online RL policy optimization is driven entirely by the reward function. Consequently, the 
interactive exploration inherent in online RL leads to the forgetting of policies acquired during IL pretraining. 

\subsection{Ablation of the Trace-Decay Parameter}
\begin{wraptable}{r}{0.5\textwidth}
    \vspace{-33pt}
    \centering
    \caption{Ablation study on the Trace-Decay parameter ($\lambda$).}
    \label{tab:ablation_trace_decay}
    \footnotesize
    \setlength\tabcolsep{2.0mm} 
    \begin{tabular}{c c c}
        \toprule
        Trace-Decay ($\lambda$) & DS (\%) $\uparrow$ & SR (\%) $\uparrow$ \\
        \midrule
        0.90 & 74.90 & 48.38 \\
        0.95 & 75.06 & 49.03 \\
        \textbf{1.00 (MC)} & \textbf{78.04} & \textbf{55.09} \\
        \bottomrule
    \end{tabular}
    \vspace{-16pt}
\end{wraptable}
We conduct an ablation study on the hyperparameter $\lambda$ of the GAE module in the PPO algorithm, as shown in the Tab.~\ref{tab:ablation_trace_decay}. The results show that the model achieves the best performance at $\lambda=1$, with absolute improvements of 3.14 and 2.94 in DS, and 6.81 and 6.06 in SR, compared to $\lambda=0.90$ and $\lambda=0.95$. When $\lambda=1.00$, GAE reduces to Monte Carlo (MC) return estimation. We argue that one possible reason for this performance difference is that MC estimation accelerates the convergence of the Value Network in long-horizon tasks with sparse rewards and deterministic termination, thus leading to better model performance.
\begin{figure}[t!]
    \centering
\includegraphics[width=0.95\textwidth]{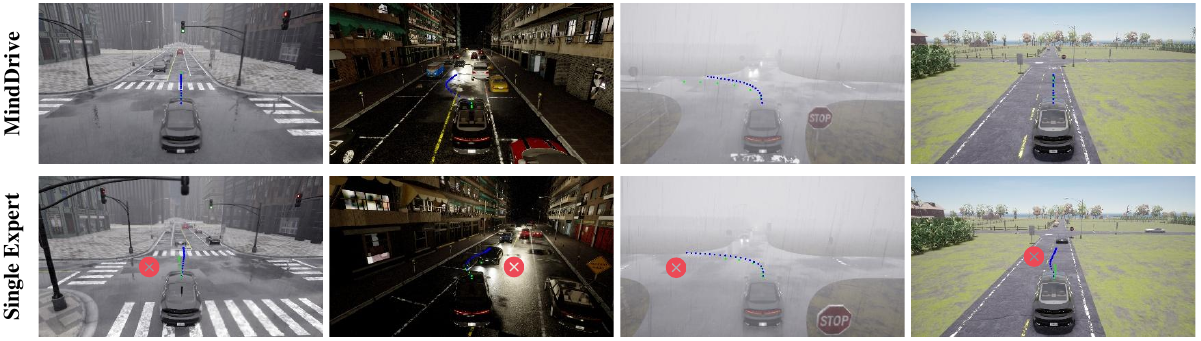}
    \caption{Comparison of our method with the single expert approach. Using a single expert results in poor trajectory quality and degraded control.}
    \label{fig:compare}
\end{figure}

\subsection{Analysis with a Single Expert}
\label{Single Expert}
To demonstrate the effectiveness of our architecture, we conduct an ablation study using a single expert to generate both trajectories and reasoning outputs during reinforcement learning. As shown in Fig.~\ref{fig:compare}, we observe that using a single expert adversely affects the trajectory output under our sparse reward setting, leading to catastrophic forgetting and a reduction in the quality of the generated trajectory. This observation also strongly validates the advantage of our framework, which decouples reasoning from action and leverages action-based rewards to guide the optimization of reasoning.

\begin{table}[t!]
\centering
\caption{Performance comparison across different numbers of training epochs. M: Merging, O: Overtaking, EB: Emergency Brake, GW: Give Way, TS: Traffic Sign, DS/SR: Driving Score/Success Rate.}
\label{tab:ab_epochs}
\footnotesize
\setlength\tabcolsep{2.4mm} 
\begin{tabular}{c cc cccccc}
\toprule
\multirow{2}{*}{Epoch} & \multicolumn{2}{c}{Closed-loop $\uparrow$} & \multicolumn{6}{c}{\textbf{Ability} (\%) $\uparrow$} \\
\cmidrule(lr){2-3} \cmidrule(lr){4-9}
& DS  & SR & {M} & {O} & {EB} & {GW} & TS & \textbf{Mean}\\
\midrule
 0 & 75.85 & 49.30 & 30.26 & 66.67 & 60.00 & 33.33 & 56.91 & 49.44 \\
 1 & 75.71 & 48.61 & 28.95 & 64.44 & 60.00 & 50.00 & 54.26 & 51.53 \\
 2 & \textbf{78.04} & \textbf{55.09} & 32.89 & \textbf{75.56}  & \textbf{68.33 } & 50.00 & 57.89 & \textbf{56.94} \\
 3 & 77.38 & 53.24 & \textbf{34.21} & 64.44 & 66.67 & 50.00 & \textbf{61.05} & 55.27  \\
 4 & 74.12 & 46.30 & 27.63 & 53.33 & 61.67 & 50.00 & 54.21 & 49.36 \\
 5 & 73.69 & 45.12 & 28.37 & 48.89 & 61.07 & 40.00 & 55.38 & 46.73 \\
\bottomrule
\end{tabular}
\end{table}

\subsection{Analysis on Performance Degradation With Online RL Training Epochs}
This section provides a detailed analysis to support the claim in the main manuscript that performance degradation with excessive training epochs stems from catastrophic forgetting. As summarized in Tab.~\ref{tab:ab_epochs}, the model's performance exhibits a clear non-monotonic relationship with the number of training epochs. The model achieves its best overall performance at 2 training epochs, with the highest mean ability score (56.94\%) and superior closed-loop metrics (DS: 78.04, SR: 55.09\%). Specifically, key capabilities such as Overtaking (75.56\%) and Emergency Brake (68.33\%) peak at this point, indicating that moderate online RL fine-tuning effectively leverages past learned knowledge while adapting to new experiences.

However, when comparing the policy at 2 epochs to configurations with more epochs, we observe a drop in the model’s ability. Early epochs allow the model to usefully incorporate new on-policy data. However, with excessive training, the policy undergoes too many updates primarily on recent interaction trajectories. This may cause it to overfit to recent experiences and forget the broader scene understanding and generalizable skills encoded during the pre-training and early fine-tuning stages. 

\subsection{More Qualitative Results}
\begin{figure}[t!]
    \centering
\includegraphics[width=0.95\textwidth]{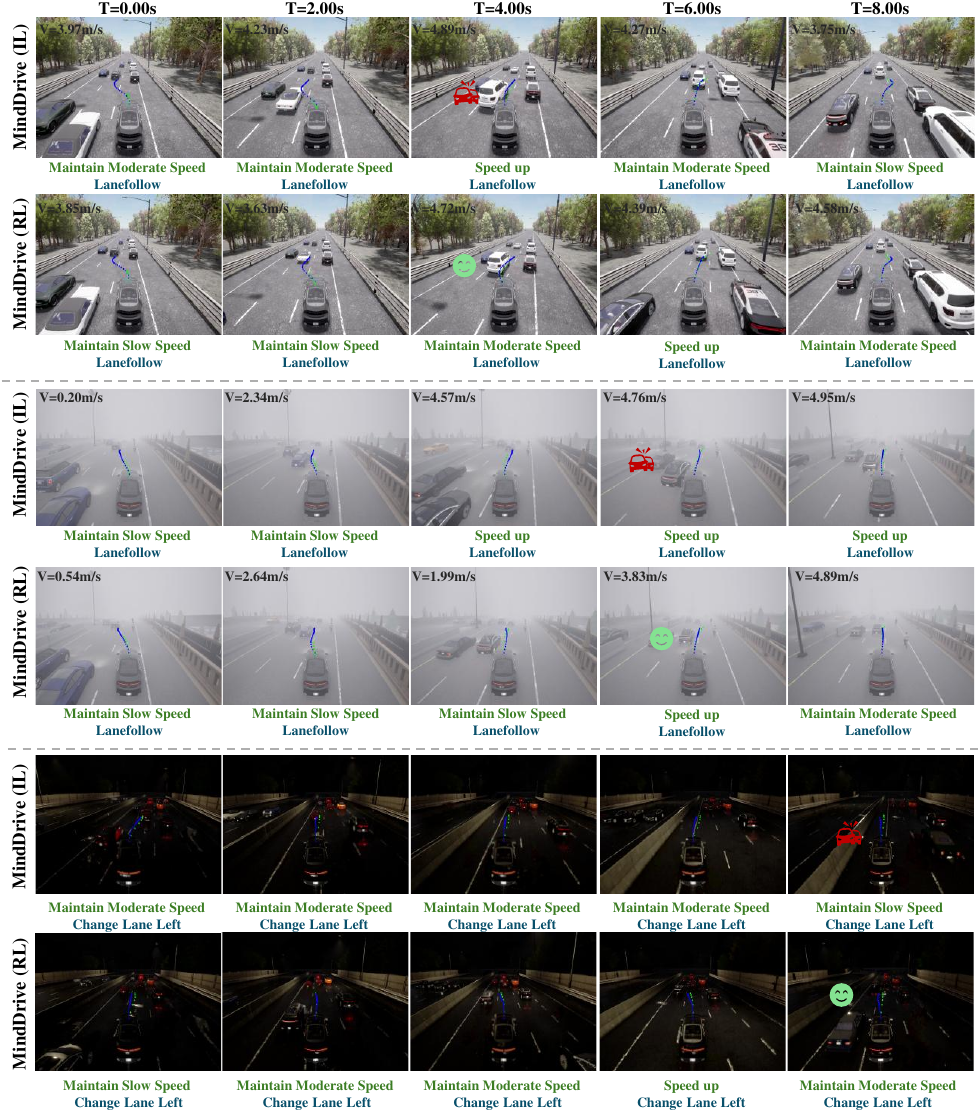}
    \caption{Comparison of our method with IL and RL.}
    \label{fig:il_rl_compare}
\end{figure}

We conduct a more comprehensive comparison between our method and imitation learning. As shown in Fig.~\ref{fig:il_rl_compare}, integrating reinforcement learning leads to a noticeable improvement in task success rate, demonstrating the effectiveness of RL-based optimization in enhancing decision-making capabilities within interactive scenarios.

\begin{figure}[t!]
    \centering
\includegraphics[width=0.95\textwidth]{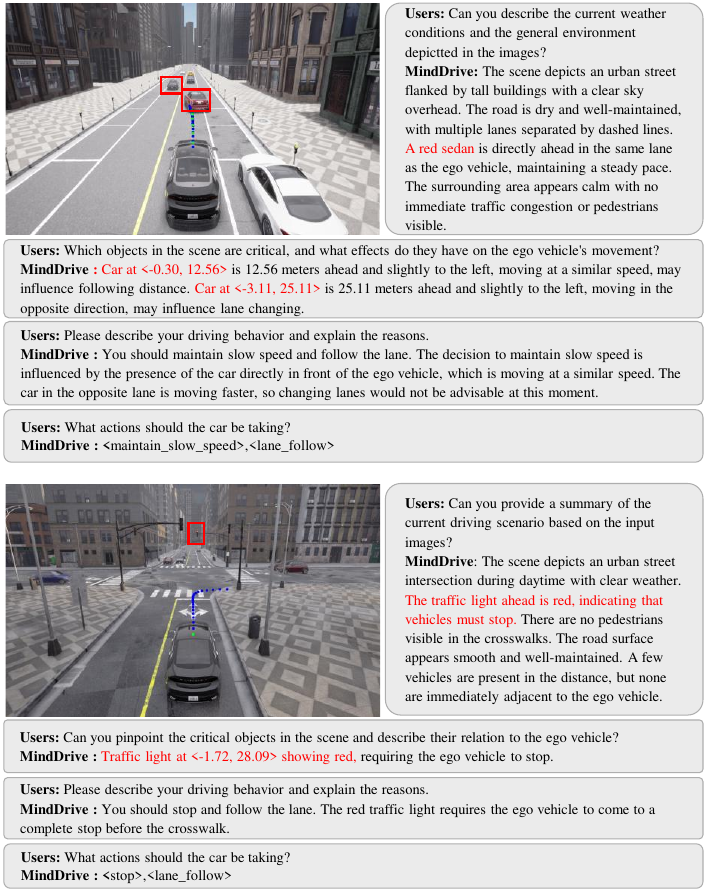}
    \caption{ Qualitative results for scene understanding and action reasoning on Bench2Drive closed-loop validation. The \textcolor{red}{red} indicates the critical objects influencing the action of the ego vehicle.}
    \label{fig:vis_qa}
\end{figure}

We also visualize the model's output during the reasoning phase, as shown in Fig.~\ref{fig:vis_qa}. Our model maintains the integrity of reasoning throughout the policy update process, demonstrating that our approach effectively performs reinforcement at the semantic level without compromising high-level cognitive functions. 




\end{document}

%% file: table/main_table.tex
\begin{table}[tp!]
\centering
\caption{Closed-loop and Multi-Ability Results of E2E-AD Methods in Bench2Drive under \textbf{base} training set. C/L refers to camera/LiDAR. * and $\dag$ denote expert feature distillation and imitation learning, respectively. "-L" represents larger model. DS: Driving Score, SR: Success Rate, M: Merging, O: Overtaking, EB: Emergency Brake, GW: Give Way, TS: Traffic Sign. Avg. L2 is averaged over the predictions in 2 seconds under 2Hz, similar to UniAD.
\label{tab: main_table}}

\footnotesize

\resizebox{\textwidth}{!}{
\setlength\tabcolsep{1.00mm} 
\begin{tabular}{l c c c c >{\columncolor{gray!10}}c >{\columncolor{gray!10}}c   ccccc >{\columncolor{gray!10}}c c} 
\toprule
\multirow{2.2}{*}{\textbf{Method}} &\multirow{2.2}{*}{\textbf{LLM}} & \multirow{2.2}{*}{\textbf{Reference}}  &\multirow{2.2}{*}{\textbf{Modality}} &\multirow{2.2}{*}{\textbf{Scheme}} & \multicolumn{2}{c}{\textbf{Closed-loop}}  & \multicolumn{6}{c}{\textbf{Ability} (\%) $\uparrow$}& \color{gray}Open-loop \\ 
\cmidrule(lr){6-7}  \cmidrule(lr){8-13}   \cmidrule(lr){14-14} & & &  & & DS$\uparrow$   & SR(\%)$\uparrow$  & {M}& {O} & {EB} & {GW} & TS& \textbf{Mean} &  \color{gray}Avg. L2 $\downarrow$ \\ 
\midrule
\midrule
\multicolumn{14}{c}{\textit{Traditional End-to-End Paradigm}} \\
\midrule

TCP-traj* & -  & NeurIPS 22 & C  &IL &  59.90     & 30.00      & 8.89       & 24.29           & 51.67       &  40.00        & 46.28    & 34.22  & \color{gray}1.70   \\
ThinkTwice*~\cite{jia2023thinktwice} & -  & CVPR 23  & C &IL   & 62.44     & 31.23     & 27.38       &  18.42          & 35.82       &  50.00        & 54.23    & 37.17    & \color{gray}0.95  \\
DriveAdapter*~\cite{jia2023driveadapter}& -  &ICCV 23  & C\&L &IL  &  64.22   & 33.08   & 28.82       &26.38           & 48.76      &  50.00         & 56.43   & 42.08    & \color{gray}1.01     \\  
UniAD-Base~\cite{hu2023planning} & -  & CVPR 23  & C &IL  & 45.81     & 16.36   & 14.10       & 17.78          & 21.67       &  10.00       & 14.21    & 15.55  & \color{gray}0.73    \\
VAD~\cite{jiang2023vad} & - & ICCV 23  & C &IL  &42.35     & 15.00    & 8.11       & 24.44         & 18.64       &  20.00       & 19.15    & 18.07 &\color{gray}0.91    \\
GenAD~\cite{zheng2024genad} & -  & ECCV 24  &C &IL & 44.81 & 15.90  &- &- &- &- &- &-&\color{gray}-\\
MomAD\cite{song2025momad} & -  & CVPR 25  & C &IL & 44.54	&16.71&- &- &- &- &- &-&\color{gray}0.87\\
WoTE~\cite{li2025end} & -  & ICCV 25 & C\&L & IL &61.71 &31.36 &- &- &- &- &- &-&\color{gray}-\\
DriveTransformer-L~\cite{jiadrivetransformer} & -  & ICLR 25   & C &IL & 63.46     & 35.01   & 17.57 & 35.00 & 48.36 & 40.00 &52.10 &38.60&  \color{gray}0.62 \\ 
DiffAD~\cite{wang2025diffad} & -  & arXiv 25 & C & IL &67.92 &38.64 &30.00 &35.55 &46.66 &40.00 &46.32 &38.79 & \color{gray}1.55\\
PGS~\cite{huang2025prioritizing} & -  & ICLR 26 & C & IL &78.08 &48.64 &35.00 &73.33 &55.00 &60.00 &43.68 &53.40 & \color{gray}0.77\\
DriveDPO~\cite{shang2025drivedpo}  & -  & NeurIPS 25 & C & Offline RL &  62.02 & 30.62 &- &- &- &- &- &- & \color{gray}- \\ 
Raw2Drive~\cite{yang2025raw2drive} &- & NeurIPS 25 &C & Online RL & 71.36 &50.24  & \textbf{43.35} &51.11 &60.00 &50.00 &62.26 &53.34 & \color{gray}-\\ 

\midrule
\multicolumn{14}{c}{\textit{Vision-Language-Action Paradigm}} \\
\midrule
Drive$\pi$0~\cite{yang2025drivemoe} & Paligemma-3B & arXiv 25 & C & IL &60.45 &30.00 &29.35 &36.58 &48.83 & 40.00 & 54.45 &  41.84 & \color{gray}0.56\\
DriveMoE~\cite{yang2025drivemoe} & Paligemma-3B & arXiv 25 & C & IL &74.22 &48.64  & 34.67 & 40.00 & 65.45 & 40.00 & 59.44 & 47.91 & \color{gray}\textbf{0.38} \\
ORION~\cite{fu2025orion} & Vicuna-7B & ICCV 25 & C & IL & 77.74 &54.62 &25.00 & 71.11 & \textbf{78.33} &30.00 &69.15 & 54.72&  \color{gray}0.68 \\
ORION~\cite{fu2025orion}  & Qwen2-0.5B & ICCV 25  & C & IL & 72.89  & 45.83 &26.32 & 62.22  & 55.55  & 50.00 &63.30 &51.37 & \color{gray}0.67\\
UniDrive-WM~\cite{xiong2026unidrive} &  Vicuna-7B & arXiv 26 & C & IL &  79.22 & 56.36  &29.81 &74.04 &79.84 &40.00 &71.30 & 59.01 & \color{gray}0.64 \\ 

RecogDrive~\cite{li2025recogdrive} &  Qwen2.5-7B & ICLR 26 & C & Offline RL &  71.36 & 45.45 &29.73 &20.00 &69.09 &20.00 &\textbf{71.34} &42.03 & \color{gray}- \\ 
MindDriver~\cite{zhang2026minddriver} &  Qwen2.5-3B & CVPR 26 & C & Offline RL &  65.48 & 39.55 & - &- &- &- &- &- & \color{gray}- \\ 
AutoVLA~\cite{zhou2025autovla} &  Qwen2.5-3B & NeurIPS 25 & C & Offline RL &  78.84 & 57.73 & - &- &- &- &- &- & \color{gray}- \\

\rowcolor[RGB]{230, 242, 255} \textbf{MindDrive{$^\dag$}}  & Qwen2-0.5B  & - & C &IL & 75.85 & 49.30 & 30.26 & 66.67 & 60.00 & 33.33 & 56.91 & 49.44 & \color{gray}0.69\\

\rowcolor[RGB]{230, 242, 255} \textbf{MindDrive-L{$^\dag$}}  & Qwen2.5-3B  & - & C &IL & 75.78 & 50.71 & 33.33 &57.78 & 68.33 & 40.00 &61.05 & 52.10 & \color{gray}0.66\\
\rowcolor[RGB]{230, 242, 255} \textbf{MindDrive} & Qwen2-0.5B & -  & C & Online RL & 78.04 & 55.09 & 32.89 & \textbf{75.56}  &68.33  & 50.00 &57.89 & 56.94 & \color{gray}0.69 \\
\rowcolor[RGB]{230, 242, 255} \textbf{MindDrive-L} & Qwen2.5-3B & -  & C & Online RL & \textbf{80.59} & \textbf{58.26} & 42.31 & 66.67  &76.67  & \textbf{50.00} &68.42 &\textbf{60.81} & \color{gray}0.66 \\
\bottomrule
\end{tabular}
}
\end{table}

%% file: table/ab_events.tex
\begin{table}[tp!]
\centering
\caption{Ablation on different penalty events. C: Collision, TL: Traffic Light, RD: Route Deviation, S: Stop. M: Merging, O: Overtaking, EB: Emergency Brake, GW: Give Way, TS: Traffic Sign, DS/SR: Driving Score/Success Rate.}
\label{tab: ab_events}

\footnotesize
\setlength\tabcolsep{1.3mm} 
\begin{tabular}{ccccc  cc ccccc >{\columncolor{gray!10}}c}
\toprule
 \multirow{2}{*}{ID} & \multicolumn{4}{c}{Penalty Events} & \multicolumn{2}{c}{Closed-loop $\uparrow$} & \multicolumn{6}{c}{\textbf{Ability} (\%) $\uparrow$}\\
\cmidrule(lr){2-5} \cmidrule(lr){6-7} \cmidrule(lr){8-13}
& C & TL & RD & S & DS  & SR(\%) & {M}& {O} & {EB} & {GW} & TS& \textbf{Mean}\\

\midrule
 1&  &  &  &  & 75.85 & 49.30 & 30.26 & 66.67 & 60.00 & 33.33 & 56.91 & 49.44 \\
 2& \checkmark &  &  &  & 75.41 & 50.70 & 28.00 & 71.11 & 60.00 & 50.00 & 56.91 & 53.20 \\
 3& \checkmark & \checkmark &  &  & 76.02  & 51.47& 31.42 & 57.77 & 68.97 & 50.00 & 58.43\textbf{ }& 53.32 \\
 4& \checkmark &  \checkmark&  \checkmark &  & 76.95 & 51.85& 27.63 & 73.33 & 65.00 & 50.00 & 57.36 & 54.67  \\
 5& \checkmark  &\checkmark  &\checkmark  &\checkmark & \textbf{78.04} & \textbf{55.09} & 32.89 & 75.56  &68.33  & 50.00 &57.89 &\textbf{56.94} \\
\bottomrule
\end{tabular}
\end{table}

%% file: table/supp/sup_route_compare.tex
\begin{table}[tp!]
    \centering
    \tiny
    \caption{Route Scenario Performance Comparison between IL and Online RL}
    \vspace{-8pt}
    \label{tab:route_performance}
    \setlength{\tabcolsep}{2.2mm}
    \renewcommand{\arraystretch}{0.9} 
    \begin{tabular}{l c c c c}
        \toprule
        Route\_id & Scenario & Skill & IL & RL \\
        \midrule
        14194 & PedestrianCrossing & Traffic Sign & 100.00 & 100.00 \\
        17563 & SequentialLaneChange & Merging & 100.00 & 100.00 \\
        17569 & SequentialLaneChange & Merging & 100.00 & 100.00 \\
        17635 & SequentialLaneChange & Merging & 34.23 & 22.69 \\
        17773 & ParkedObstacle & Overtaking & 60.00 & 100.00 \\
        1792 & HazardAtSideLane & Overtaking & 100.00 & 100.00 \\
        1656 & ParkingExit & Merging & 60.00 & 60.00 \\
        2164 & VehicleTurningRoutePedestrian & Emergency Brake & 25.21 & 80.00 \\
        2201 & EnterActorFlow & Traffic Sign & 25.20 & 36.00 \\
        23670 & HighwayExit & Merging & 100.00 & 100.00 \\
        2373 & VanillaSignalizedTurnEncounterRedLight & Traffic Sign & 70.00 & 70.00 \\
        23901 & InterurbanActorFlow & Merging & 100.00 & 100.00 \\
        23930 & InterurbanActorFlow & Merging & 97.24 & 97.24 \\
        2397 & VanillaNonSignalizedTurn & Traffic Sign & 65.00 & 100.00 \\
        24041 & HighwayExit & Merging & 100.00 & 100.00 \\
        24071 & InterurbanAdvancedActorFlow & Merging & 60.00 & 60.00 \\
        24098 & InterurbanActorFlow & Merging & 94.12 & 94.12 \\
        25300 & HazardAtSideLane & Overtaking & 100.00 & 100.00 \\
        2534 & Accident & Overtaking & 100.00 & 100.00 \\
        25439 & HazardAtSideLane & Overtaking & 100.00 & 100.00 \\
        2554 & ParkedObstacle & Overtaking & 100.00 & 60.00 \\
        25854 & hazardAtSideLaneTwoWay & Overtaking & 60.00 & 60.00 \\
        25863 & PedestrianCrossing & Overtaking & 70.00 & 100.00 \\
        2606 & ConstructionObstacleTwoWays & Emergency Brake & 100.00 & 100.00 \\
        26950 & OppositeVehicleRunningRedLight & Overtaking & 100.00 & 100.00 \\
        27018 & SignalizedJunctionRightTurn & Emergency Brake & 60.00 & 60.00 \\
        27529 & PedestrianCrossing & Traffic Sign & 80.00 & 60.00 \\
        28085 & T\_junction & Emergency Brake & 70.00 & 70.00 \\
        28098 & T\_junction & Traffic Sign & 70.00 & 70.00 \\
        28049 & SignalizedJunctionLeftTurnEnterFlow & Traffic Sign & 60.00 & 70.00 \\
        28210 & NonSignalizedJunctionLeftTurnEnterFlow & Traffic Sign & 60.00 & 36.00 \\
        2898 & BlockedIntersection & Merging & 20.93 & 60.00 \\
        3178 & VanillaNonSignalizedTurn & Emergency Brake & 65.00 & 65.00 \\
        3307 & Accident & Traffic Sign & 60.00 & 100.00 \\
        3380 & YieldToEmergVehicle & Overtaking & 70.00 & 70.00 \\
        3410 & AccidentTwoWays & Give Way & 60.00 & 60.00 \\
        3514 & ParkingExit & Overtaking & 60.00 & 100.00 \\
        3714 & VehicleTurningRoutePedestrian & Merging & 30.52 & 100.00 \\
        3785 & MergelntoSloTraffic & Emergency Brake & 100.00 & 89.32 \\
        3876 & VanillaSignalizedTurnEncounterRedLight & Merging & 70.00 & 70.00 \\
        3904 & VanillaNonSignalizedTurnEncounterStopsign & Traffic Sign & 100.00 & 80.00 \\
        3936 & SignalizedJunctionLeftTurn& Traffic Sign & 60.00 & 60.00 \\
        4468 & SignalizedJunctionLeftTurn & Merging & 100.00 & 100.00 \\
        4683 & SignalizedJunctionLeftTurn & Merging & 60.00 & 60.00 \\
        \bottomrule
    \end{tabular}
\end{table}